\documentclass[conference,letterpaper]{IEEEtran}
\IEEEoverridecommandlockouts

\usepackage{cite}
\usepackage{amsmath,amssymb,amsfonts}
\usepackage{algorithmic}
\usepackage{graphicx}
\usepackage{textcomp}
\usepackage{xcolor}
\usepackage{mathtools} 
\usepackage{mathrsfs}
\usepackage{graphicx}
\usepackage{multirow}
\usepackage{hyperref}

\newcommand{\dy}{\mathsf{d}}
\newcommand{\h}{\mathsf{h}}
\newcommand{\m}{\mathsf{m}}
\newcommand{\w}{\mathsf{w}}

\newcommand{\ACA}{\mathrm{ACA}}
\newcommand{\att}{\mathrm{Att}}

\newcommand{\ReLU}{\mathrm{ReLU}}

\usepackage{lipsum}

\def\BibTeX{{\rm B\kern-.05em{\sc i\kern-.025em b}\kern-.08em
    T\kern-.1667em\lower.7ex\hbox{E}\kern-.125emX}}

\makeatletter
\let\@fnsymbol\@arabic
\makeatother
\begin{document}

\title{GACAN: Graph Attention-Convolution-Attention Networks for Traffic Forecasting Based on Multi-granularity Time Series{\LARGE \textsuperscript{*}}
	\thanks{*\:This paper has been published in the Proceedings of 2021 International Joint Conference on Neural Networks (IJCNN 2021) \tt \url{https://ieeexplore.ieee.org/document/9534064/}}
}

\author{\IEEEauthorblockN{
		Sikai Zhang\IEEEauthorrefmark{2}, Hong Zheng\IEEEauthorrefmark{2}, 
		Hongyi Su\IEEEauthorrefmark{2}, Bo Yan\IEEEauthorrefmark{2}, Jiamou Liu\IEEEauthorrefmark{1}, Song Yang\IEEEauthorrefmark{1}}

		\IEEEauthorblockA{\IEEEauthorrefmark{2}Beijing Institute of Technology, China}
		\IEEEauthorblockA{\IEEEauthorrefmark{1}The University of Auckland,Auckland, New Zealand
		}
		hongzheng@bit.edu.cn, jiamou.liu@auckland.ac.nz, song.yang@auckland.ac.nz
}

%

\maketitle


\begin{abstract}
Traffic forecasting is an integral part of intelligent transportation systems (ITS). Achieving a high prediction accuracy is a challenging task due to a high level of dynamics and complex spatial-temporal dependency of road networks. For this task, we propose Graph Attention-Convolution-Attention Networks (GACAN). 
The model uses a novel Att-Conv-Att (ACA) block which contains two graph attention layers and one spectral-based GCN layer sandwiched in between. The graph attention layers are meant to capture temporal features while the spectral-based GCN layer is meant to capture spatial features. The main novelty of the model is the integration of time series of four different time granularities: the original time series, together with hourly, daily, and weekly time series. Unlike previous work that used multi-granularity time series by handling every time series separately, GACAN combines the outcome of processing all time series after each graph attention layer. Thus, the effects of different time granularities are integrated throughout the model. We perform a series of experiments on three real-world datasets. The experimental results verify the advantage of using multi-granularity time series and that the proposed GACAN model outperforms the state-of-the-art baselines.  
\end{abstract}


\section{Introduction}
Forecasting traffic conditions has long been an issue of great interest. According to a 2018 survey ({\small \url{https://aaafoundation.org/american-driving-survey-2015-2016}}), drivers in the U.S. spend 50.6 minutes on the road and drive 31.5 miles per day on average. To effectively mitigate the increasing traffic load and alleviate potential congestion, there has been intensive studies on designing next-generation traffic management systems. In particular, intelligent transportation systems offer a suite of tools that provide advanced technologies for the measurement, prediction, and control of traffic environment \cite{sumalee2018smarter}. By a traffic environment, we mean a road network that consists of locations and roads that connect these location. As sensing devices such as cameras and sensors are increasingly used to measure traffic conditions, the amount of traffic data is growing at an unprecedented rate. The growth in available data has triggered considerable interests on ITS. As an integral component of any ITS, the task of {\em traffic forecasting} is important to numerous functionalities such as traffic speed control, route planning and navigation \cite{yu2018spatio}.  

 Traffic forecasting aims to predict the value of key traffic indicators, such as vehicle speed, traffic volume, and density, at specified locations in the traffic network, based on historical sensor readings of these indicators. The input to the problem consists of time series gathered at sensing devices across the network: Nodes in the network represent sensor locations and edges represent road segments connecting the sensor locations. Sensor readings are represented as attributes on nodes  which are  changing with time. This network provides the foundation for a data-rich solution to the traffic forecasting problem. With this setup, the task may exploit two dimensions of information extraction: The first is the {\em temporal} dimension as historical traffic conditions will undoubtedly reveal much information regarding future conditions. The second is the {\em spatial} dimension as the traffic condition at a location may affect the condition at another location on the network. The challenge is to identify the hidden influence between locations in the network across different time. Thus traffic forecasting amounts to a typical spatial-temporal data mining task. We use a very simple example to illustrate the challenges we face in this task; see Figure~\ref{fig:fig1}. The example shows a simple network where the traffic conditions of locations affect each other. In particular, two locations exhibit different mutual effects across the same time on different days (as shown in (a)), and across different hours on the same day (as shown in (b)). In the following paragraphs we briefly summarize challenges faced by traffic forecasting. 
 
 \begin{figure*}
    \centering
    \includegraphics[width=0.9\textwidth]{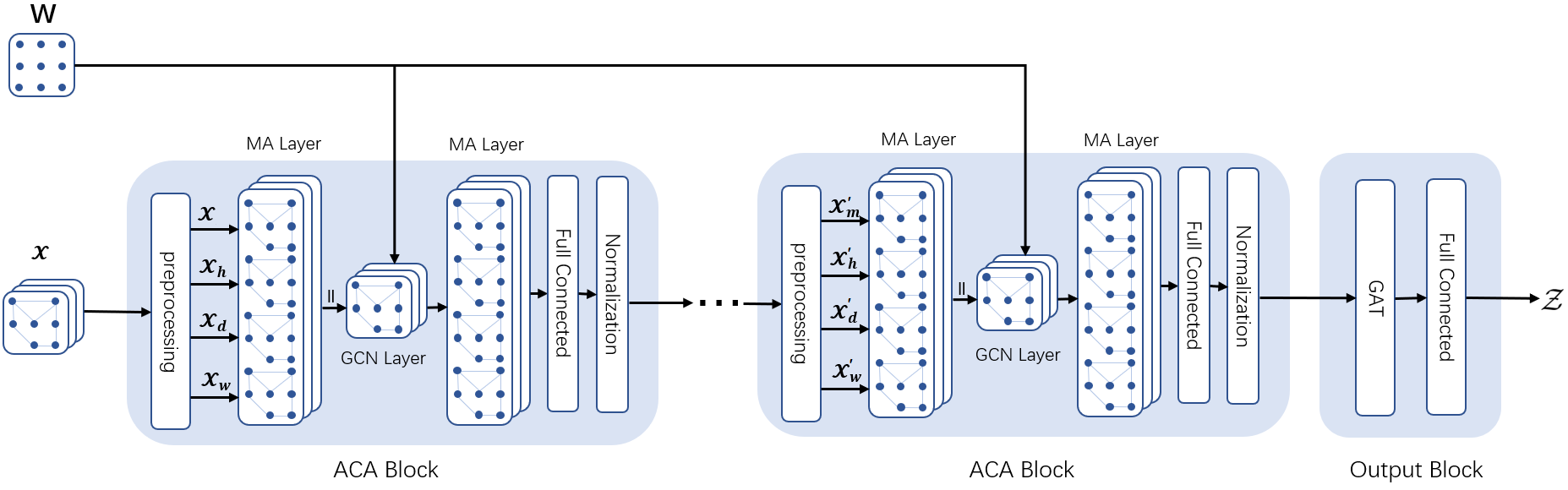}
    \caption{The architecture of GACAN. GACAN consist of several att-conv-att (ACA) blocks and finally a single-head graph attention with a fully-connected output layer. }
    \label{fig:fig2}
\end{figure*}
 
\begin{figure}[!htbp]
    \centering
    \includegraphics[width=0.44\textwidth]{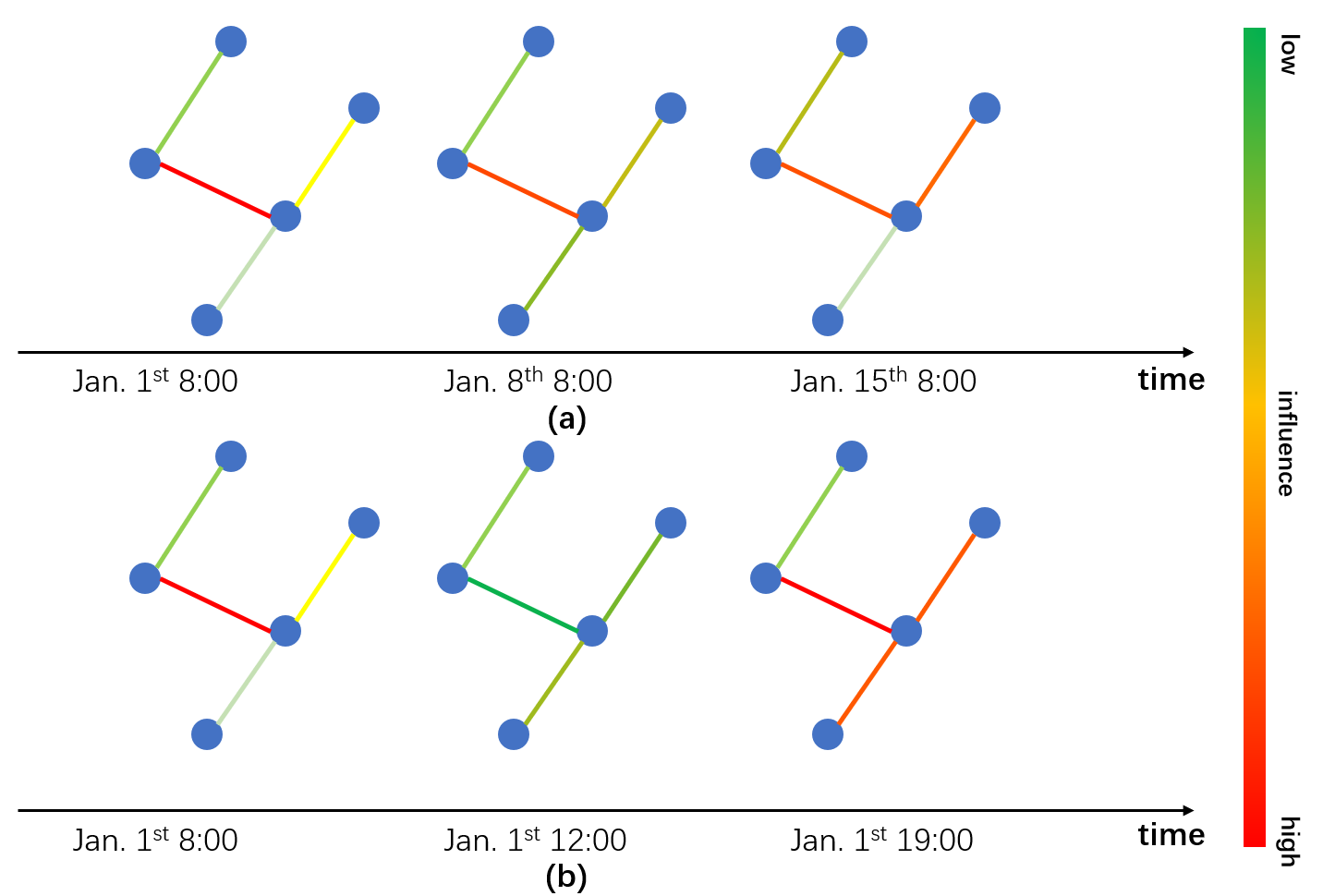}
    \caption{The nodes (shown in blue) represent sensor locations. The color of an edge represents the level of correlation between the two endpoints. In the spatial dimension, traffic conditions at different locations may affect a location differently. In the temporal dimension, the correlation between two locations may change across different time granularities.}
    \label{fig:fig1}
\end{figure}

\paragraph*{\bf 1. Short-term v.s. medium-and-long term prediction}  In terms of the time horizon of prediction, traffic forecasting tasks can be broadly classified as {\em short-term} traffic prediction with horizon ranging from a few minutes to half an hour, or {\em medium-and-long term} traffic prediction where the horizon is larger than 30 minutes. Traditionally, autoregressive models such as moving average and vector autoregressive model (VAR) are widely applied for time series prediction. However, these methods mostly rely on the linearity assumption made on the time series dataset. The dynamics of traffic conditions, on the other hand, is highly non-linear, thus presenting severe limitations when these methods are applied to medium-and-long term traffic prediction tasks. 

\paragraph*{\bf 2. Dynamical v.s. data-driven modeling} Past studies on medium-and-long term traffic prediction generally fall into two categories: {\em dynamical modeling}\cite{vlahogianni2015computational} and {\em data-driven modeling}\cite{yu2018spatio}. The former makes use of sophisticated simulation paradigms such as agent-based modeling and ant-colony optimization 
to resemble crowd-dynamics from a {\em microscopic} perspective \cite{doolan2014time}. 
Due to the complexity, instability and interference of traffic conditions, these models tend to depart considerably from real-life scenarios. The latter, on the other hand, aims to discover patterns of network dynamics directly from data without incorporating domain knowledge. These models are usually {\em macroscopic} in the sense that they directly output the states of the overall traffic speed, rather than of individuals in the system. We focus on the second category. Along this line of research, neural-based models, such as the ones that are based on convolutional neural networks (CNN), are gaining popularity in recently years \cite{zhang2016dnn,zhang2017deep}.

\paragraph*{\bf 3. Spatial v.s. temporal feature extraction} More recently, graph convolutional networks (GCN) are adopted to better capture spatial and temporal features in a traffic network \cite{yu2018spatio,li2018diffusion,guo2019attention}. Two types of GCN have been used, {\em spectral-based}  and {\em spatial-based}. The former is defined using graph Fourier transform  \cite{bruna2013spectral}, which projects an input graph signal to the orthonormal space whose basis consists of eigenvectors of the normalized graph Laplacian of the traffic network. Due to the use of filters, graph convolution may be interpreted as a denoising operation. Existing work, e.g., the STGCN model defined in \cite{yu2018spatio}, have used spectral-based GCN to capture spatial features of a traffic network. The latter type treats graph convolution as an operation that updates a node's representation by aggregating those of its neighbors. One notable example of a spatial-based GCN is the graph attention network (GAT) \cite{velivckovic2017graph} which is a variant of the self-attention mechanism. These models are natural candidates to use when the goal is to extract temporal correlations in traffic forecasting.

\paragraph*{\bf 4. Time granularity} 
The sampling frequency of road network sensors are sufficiently high to capture fine-grained temporal information regarding traffic speed dynamics, e.g., Caltrans PeMS measures traffic once every 5 minutes \cite{chen2001freeway}. However, when predicting traffic speed, it is sometimes beneficial to consider time series that are coarser-grained. This is largely due to periodicity of the traffic conditions, which are persistent patterns on a daily or weekly basis. For example,  Guo et. al.\cite{guo2019attention} utilized three time series when training  their ASTGCN model, where values are separated by 5 minutes, one day, and one week, respectively \cite{guo2019attention}. While the 5-minute values show traffic dynamics immediately before the to-be-predicted time, the daily values are indicative of daily flow patterns such as peak-hour congestion, and the weekly values reflect patterns such as shifts between mid-week and weekend.  While ASTGCN demonstrated superior prediction accuracy compared to other models, the three time series are processed separately, resulting in three different spatial-temporal features \cite{guo2019attention}. This leaves open the possibility that certain latent spatial-temporal features are better expressed when multiple time granularities are combined during feature extraction. It is thus a question how time series data of different time granularity may be integrated for better outcome.   

\paragraph*{\bf Contribution} Capitalizing on recent progress on GCN, we proposed {\em \textbf{G}raph \textbf{A}ttention-\textbf{C}onvolutional-\textbf{A}ttention \textbf{N}etwork} (GACAN), a novel data-driven model for traffic forecasting. The novelty of the model is two-fold: First, we combine both spectral-based and spatial-based GCN by  inventing the Att-Conv-Att (ACA) block. An ACA block contains two graph attention layers and one spectral-based GCN layer sandwiched in between. In this way, we take advantage of graph attention's ability to extract temporal correlations and spectral-based GCN's ability to capture spatial correlations. The block extracts temporal feature from input signal (via graph attention), which then affect the extraction of spatial features (via spectral-based GCN), which in turn helps to refine the temporal feature extraction (via graph attention again). Second, inspired by existing method that utilize time series of multiple time granularities, we process hourly, daily and weekly traffic speed values on top of the original time series input. Unlike the ASTGCN model \cite{guo2019attention}, the outcomes of processing inputs of different time granularities are fused after each graph attention layer, hence the extracted spatial-temporal features are affected by all time series. 

Since all components of GACAN are convolutional structures, the model drastically reduces training time as compared with models that are based on recurrent structures. The combination of spectral-based GCN and graph attention based on multi-granularity time series helps to achieve high accuracy. We validate GACAN using a series of experiments over three real-world traffic forecasting datasets with prediction horizon up to an hour. The experiments verify that our framework outperforms existing baselines, i.e. compared with a SOTA model ASTGCN, the average performance of GACAN is improved by 5.2\% in the tested scenarios.

\section{Related Work}

Accurate traffic forecasting has been a challenging problem since the early 2000s and many methods have been proposed. Early models such as auto-regressive integrated moving average (ARIMA) \cite{williams2003modeling} and vector auto-regression (VAR) \cite{zivot2006vector} utilize statistical analysis methods that mainly focus on the static assessment of time series and neglect the temporal-spatial dependency. It is thus unsurprising that the accuracy of these models,  when applied to medium-and-long term traffic forecasting, is severely limited. Many machine learning algorithms such as k-nearest neighbors algorithm (KNN) \cite{van2012short} and support vector machine (SVM) \cite{jeong2013supervised} are adopted in the early 2000s with improved accuracy. 

In the last 3-4 years, 
emphasis has been shifted to neural-based models for traffic forecasting, with the adoption of, e.g., deep belief network (DBN) \cite{jia2016traffic} and stacked autoencoder (SAE) \cite{lv2014traffic,chen2016learning}. 
To extract spatial-temporal features jointly, a number of spatial-temporal deep learning models are proposed: Wu {\em et al.} \cite{wu2016short} combined CNN and LSTM to align spatial and temporal regularities,  Zhang {\em et al.} \cite{zhang2018predicting} proposed ST-ResNet that uses residual convolutional units to model the temporal closeness, period, and trend properties of crowd traffic. Wang {\em et al.}  \cite{wang2018cross} proposed RegionTrans to transfer knowledge from a data-rich source city to a data-scarce target city. Yao {\em et al.} \cite{yao2018modeling} in spatial-temporal dynamic network (STDN) used a flow gating mechanism and a periodically shifted attention mechanism to predict taxi demand between two similar cities. These models extract the spatial features only from grid layout, failing to capture and utilize the latent spatial-temporal features of the road network. 


To distill spatial-temporal features from graph-based data, GCN is used to develop predictive models. We next review spatial-based \cite{scarselli2008graph,gao2018large} and spectral-based GCN \cite{bruna2013spectral,li2018adaptive}.

\paragraph*{\bf Spectral-based GCN} These methods \cite{bruna2013spectral} define graph convolution using filters from the perspective of graph signal processing, thus viewing it as a denoising operation on graph signals. More specifically, spectral-based GCN is defined using graph Fourier transform \cite{gadde2013bilateral}: Given an edge-weighted network $G$ with $N$ nodes whose edge weights are represented by the $N\times N$ adjacency matrix $\mathbf{W}$. Let $D \in \mathbb{R}^{N \times N}$
be the diagonal degree matrix with $D_{i i}\coloneqq\sum_{j} \mathbf{W}_{i,j}$. The {\em normalized graph Laplacian} is defined as 
\begin{equation}\label{equ:laplacian}
    L\coloneqq I_{N}-D^{-\frac{1}{2}} \mathbf{W} D^{-\frac{1}{2}} \in \mathbb{R}^{N \times N}
\end{equation}
where $I_N \in \mathbb{R}^{N \times N}$ is the identity matrix.

Let $\lambda_1,\ldots,\lambda_N$ denote the eigenvalues of $L$ and let $\Lambda \in \mathbb{R}^{N \times N}$ be the diagonal degree matrix with $\Lambda_{i i}\coloneqq \lambda_i$. The matrix $L$ can be rewritten as $L\coloneqq U \Lambda U^{\mathsf{T}}$ where the {\em graph Fourier basis} $U \in \mathbb{R}^{N \times N}$ is the matrix of eigenvectors of $L$ corresponding to $\lambda_1,\ldots,\lambda_N$, respectively. The {\em graph convolution} of the input signal $X\in \mathbb{R}^{N\times Q}$ (for some integer $Q\geq 1$) with a kernel $\Theta\colon \mathbb{R}^{N\times N}\to \mathbb{R}^{N\times N}$ is defined as
\begin{equation}
    \Theta *_G X\coloneqq\Theta(L) X\coloneqq\Theta\left(U \Lambda U^{\mathsf{T}}\right) X=U \Theta(\Lambda) U^{\mathsf{T}} X.
    \label{equ:graph_conv_1}
\end{equation}
By this definition, a graph signal $X$ is filtered by a kernel $\Theta$ with multiplication between $\Theta$ and graph Fourier transform $U^{\mathsf{T}} X$ \cite{shuman2013emerging}, which gives rise to a convolutional network structure. The spectral-based GCN has been used to extract spatial feature of a traffic network for traffic prediction. Li {\em et al.} proposed the diffusion convolutional recurrent neural networks (DCRNN) \cite{li2018diffusion} for this task, which  takes the direction of graph edges into account. Yu {\em et al.} first combined spatial and temporal features to predict traffic speed using their STGCN model \cite{yu2018spatio}. Our GACAN model will use an efficient approximation of spectral-based GCN (see Sec.~\ref{sec:convolution}) to capture spatial features. 


\paragraph*{\bf Spatial-based GCN} The spatial-based GCN methods \cite{micheli2009neural}, on the other hand, represent graph convolution as the aggregation of feature information from the neighborhoods of nodes. Thus,  models in this category can be viewed as a type of CNN where every node's representation is updated by convolving with representations of its neighbors’.  Most notably, {\em graph attention networks} is an spatial-based method introduced in \cite{velivckovic2017graph}. The attention mechanisms \cite{vaswani2017attention}, first proposed in natural-language process to solve machine translation tasks, 
 has the ability to reveal the most important regions in the input. Hence the mechanism is especially suitable for discovering dependencies among parts of an sequential input.  
 GAT is a variant of the self-attention mechanism and is designed to handle graph input. The model aggregates neighborhood features of nodes and is thus a spatial-based GCN. The ability to reveal time-dependency in the input data makes the model especially suitable to uncover temporal features. 
  A multi-headed variant of GAT, namely {\em Multi-GAT} structure (MA), is invented to stabilize the learning process \cite{velivckovic2017graph}. MA aims to learn attention weights in multiple subspaces. More formally, let $\mathcal{N}_i$ denote the neighborhood of a node $i$ in the graph. 
 For a positive integer $K$, the $K$-headed MA is deﬁned in \eqref{equ:ma}. The input to this layer are node features $\mathbf{h}_1,\ldots,\mathbf{h}_n$ and the output are new node features $\mathbf{h}'_1,\ldots,\mathbf{h}'_n$, defined by
\begin{equation}
    \mathbf{h}'_{i} \coloneqq \|_{k=1}^{K} \sigma\left(\sum_{j \in \mathcal{N}_{i}} \alpha_{i,j}^{k} W_{k} \mathbf{h}_{j}\right)
    \label{equ:ma}
\end{equation}
where $\|$ denotes the concatenation operation, $\sigma$ is a non-linear mapping, $\alpha_{i,j}^{k}$ is the $k$th attention coefficients which adaptively controls the contribution of a neighbor $j$ to the node $i$, and $W_k$ is the learnable linear transformation that corresponds to the $k$th attention mechanism. Zhang {\em et al.} put forward a gated attention model GaAN \cite{zhang2018gaan} which uses a sub-network to control each attention head’s importance to solve node classification problems. 
For traffic forecasting, Guo {\em et al.} proposed the ASTGCN model \cite{guo2019attention} which makes use of MA to extract latent traffic features. In comparison, our GACAN model will use a variant of MA (see Sec.~\ref{sec:attention}) to capture temporal features. 

\section{Proposed Model}
\subsection{Problem Formulation}
\index A {\em traffic network} is an undirected graph $G = (V, E)$  where $V$ is a set of $|V| = N$ nodes, and the set of edges $E$ indicates local road connectivity between nodes. 
We assume a discrete time model $t=1,2,\ldots$ where time slices are separated by regular intervals. 
At any timestamp $t$, 
every node $i\in V$ has a {\em traffic speed} $\mathcal{X}_{i}^{t}$ that reflects the average vehicle speed at this node. Let $\mathcal{X}^t$ denote the speed vector $\left(\mathcal{X}_{1}^{t},\ldots,\mathcal{X}_{N}^{t}\right)$. For a {\em time horizon}  $H$, the {\em traffic forecasting problem} aims to predict the most likely traffic speed in the next $H$ time slices given the observed speed of the $Q$ preceding time slices. More formally, the problem asks for the prediction vectors at time step $t_0+1,\ldots,t_0+H$:
\begin{multline}
\hat{\mathcal{X}}^{t_0+1}, \dots, \hat{\mathcal{X}}^{t_0+H} \coloneqq \mathop{\arg\max}_{\mathcal{X}^{t_0+1} , \ldots, \mathcal{X}^{t_0+H}} \\\ln{\Pr\left[\mathcal{X}^{t_0+1} , \ldots,
\mathcal{X}^{t_0+H} | \mathcal{X}^{t_0-Q+1} , \ldots, \mathcal{X}^{t_0} \right]}
\end{multline}
where $t_0$ is the current time slice. 

It is worthy pointing out that {\em time granularity} of the input time series determines the number of traffic readings during a fixed amount of time. As discussed above, time granularity plays an important role in affecting the prediction performance of a trained model. In most real-world datasets, the interval between consecutive sensor readings is very small. For example, in the PeMS and METR datasets to be used in Sec.~\ref{sec:experiment}, the traffic speed are sampled every 5 minutes. Such fine time granularity is suitable when predicting future traffic in a short horizon. However, for medium-and-long term prediction, using  5-minute input data may distract the model away from longer term shifts in traffic conditions. For example, it can be argued that traffic conditions exhibit a clear daily patterns with slow speed and high volume during the peak hours, and much faster speed and sparse volume during after hours. 

\subsection{Network Architecture}
We now present the architecture of our proposed GACAN model. The models that bear the closest resemblance to our GACAN model are STGCN \cite{yu2018spatio} and ASTGCN \cite{guo2019attention}. 
In STGCN,  a ``sandwich'' structure was adopted to design a stackable {\em ST-Conv block}: The input is first passed into a gated CNN layer that extracts temporal features, which is then fed into a spectral-based GCN layer that extracts spatial features, before finally entering another gated CNN layer that extracts temporal feature. This structure facilitates bottleneck in the network to achieve scale compression and feature squeezing.  Moreover, layer normalization is used to prevent overfitting. On the other hand, ASTGCN capitalizes on the power of graph attention mechanism to  reveal relative importance of data to the task. The identified importance indices are then processed by a spectral-based GCN to perform traffic prediction. 


Inspired by these two earlier models, our proposed GACAN model utilizes GCN in the following way. The network is composed of a number of stackable {\em Att-Conv-Att (ACA) blocks} that capture spatio-temporal convolutional features. Similar to STGCN, each ACA block is also a ``sandwich'' structure consisting of three layers to capture temporal, spatial, and temporal features, respectively. Similar to ASTGCN, the two layers that we used for the temporal features are graph attentions (i.e., MA). 
The details of the model architecture are described as Figure~\ref{fig:fig2}. 

The input to the first ACA block is a feature sequence $\mathcal{X}\coloneqq (\mathcal{X}^{t-Q+1},\mathcal{X}^{t-Q+2},\ldots,\mathcal{X}^{t})$ (for some $t$) as defined above. 
An output layer is attached after the last ACA block which consists of an attention layer and a fully-connected layer. The attention layer maps the last ACA block's output to the prediction of $H$ future time slices. The output of GACAN with two ACA blocks $\mathcal{Z} \in \mathbb{R}^{H\times N}$ is the speed prediction for the $N$ nodes in the network and is computed by 
\begin{equation}
    \mathcal{Z}=\mathrm{FC}\circ {\mathrm{GAT}} \circ {\ACA} \circ {\ACA}(\mathcal{X})
\end{equation}
where $\mathrm{GAT}$ is a single-head graph attention (i.e., $K=1$), $\mathrm{FC}$ is a fully-connected layer with leakyReLU activation, and $\circ$ denotes function composition. We will now describe the ACA block in detail. 

\subsection{Att-Conv-Att (ACA) Block}
The ACA block is constructed to fuse spatial and temporal features from graph-based time series. An ACA block can be stacked when dealing with more complex or certain particular cases. To emphasize the importance of time granularity and improve prediction accuracy, we pre-process the input data $\mathcal{X}$ to extract three new time series inputs: the hourly, daily, and weekly traffic readings, respectively. Suppose $p$ is the number of minutes between two consecutive time slices in the {\em original} time series. Let $s_\h,s_\dy,s_\w$ be $60/p$, $24s_\h$, and $7s_\dy$, respectively, denoting the number of time slices taken in an hour, a day, and a week. Then let $t_\h,t_\dy,t_\w$ be $Q/s_\h, Q/s_\dy, Q/s_\w$, respectively. Define the input time series

\begin{equation}
\begin{aligned}
\mathcal{X}_\h \coloneqq \{&\mathcal{X}^{t_{0}-t_{\h}s_{\h}+1},\cdots,\mathcal{X}^{t_{0}-t_{\h}s_{\h}+H},\mathcal{X}^{t_{0}-(t_{\h}-1)s_{\h}+1},\cdots,\\ &\mathcal{X}^{t_{0}-(t_{\h}-1)s_{\h}+H},\cdots,\mathcal{X}^{t_{0}-s_{\h}+H}\} \\
\mathcal{X}_\dy \coloneqq \{&\mathcal{X}^{t_{0}-t_{\dy}s_{\dy}+1},\cdots,\mathcal{X}^{t_{0}-t_{\dy}s_{\dy}+H},\mathcal{X}^{t_{0}-(t_{\dy}-1)s_{\dy}+1},\cdots,\\ &\mathcal{X}^{t_{0}-(t_{\dy}-1)s_{\dy}+H},\cdots,\mathcal{X}^{t_{0}-s_{\dy}+H}\}\\
\mathcal{X}_\w \coloneqq \{&\mathcal{X}^{t_{0}-t_{\w}s_{\w}+1},\cdots,\mathcal{X}^{t_{0}-t_{\w}s_{\w}+H},\mathcal{X}^{t_{0}-(t_{\w}-1)s_{\w}+1},\cdots,\\ &\mathcal{X}^{t_{0}-(t_{\w}-1)s_{\w}+H},\cdots,\mathcal{X}^{t_{0}-s_{\w}+H}\}
\end{aligned}
\end{equation}

Namely, $\mathcal{X}_\h$, $\mathcal{X}_\dy$, $\mathcal{X}_\w$ are respectively the hourly, daily, and weekly time series. These four input time series will be used together to make the prediction $\hat{\mathcal{X}}^{t+1}, \ldots, \hat{\mathcal{X}}^{t+H}$.

\begin{figure}
    \centering
    \includegraphics[width=0.48\textwidth]{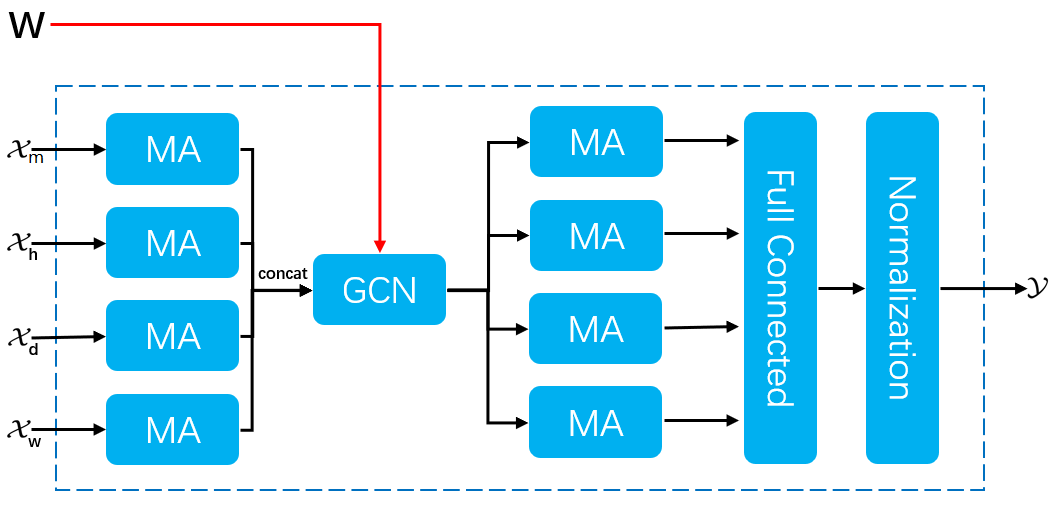}
    \caption{The architecture of an ACA block (after pre-processing). The input are the features corresponding to the four input time series: original, hourly, daily, and weekly features $\mathcal{X}_\m, \mathcal{X}_\h, \mathcal{X}_\dy, \mathcal{X}_\w$. The block contains two sets of MA layers and a GCN layer in between. Each sets of MA layers contains four independent MA for processing the four feature maps. Full connected layer combines the four MA layers' features. To prevent overfitting, a normalization layer is applied. The output is a (single) feature map $\mathcal{Y}$.}
    \label{fig:fig3}
\end{figure}
Figure~\ref{fig:fig3} displays the architecture of the ACA block after input pre-processing. There are three main components: two MA layers with the same structure and a spectral-based GCN layer in between. To prevent overfitting, each ACA block utilizes a normalization layer. 
Let $\overline{\mathcal{X}}\coloneqq (\mathcal{X},\mathcal{X}_\h,\mathcal{X}_\dy, \mathcal{X}_\w)$.
The output $\mathcal{Y}\coloneqq \ACA(\overline{\mathcal{X}})\in \mathbb{R}^{N\times P}$ of the ACA block is a $N\times P$ matrix, which is computed by
\begin{equation}
    \ACA(\mathcal{X}) \coloneqq \mathrm{norm}\circ \mathrm{FC} \circ \att \circ \ReLU (\Theta \ast_{G} {\att}(\overline{\mathcal{X}}))
\end{equation}
where $\att$ is our four independent multi-attention functions as defined in Section~\ref{sec:attention}, $\Theta$ is the spectral kernel of graph convolution as defined in \ref{sec:convolution},  $\ReLU(\cdot)$ denotes the rectified linear units function, and $\mathrm{norm}$ is the normalization function.

\subsection{Multi-Attention Block for Extracting Temporal Features}\label{sec:attention}
As Figure~\ref{fig:fig3} shows, an ACA block uses four independent MA with the same structure to capture the original, hourly, daily, and weekly feature dependencies respectively. Recall that $\mathcal{X}_\h,\mathcal{X}_\dy,\mathcal{X}_w$ denote the hourly, daily, and weekly input, respectively. Let $\mathcal{X}_\m$ denote the input in the original time granularity\footnote{For simplicity, we only define the first MA block layer whose inputs are $\mathcal{X}_\sharp=(X^{t-t_\sharp s_\sharp},X^{t-(t_\sharp-1) s_\sharp},\ldots,X^t)$ where $\sharp\in \{\m,\h,\dy,\w\}$. All MA blocks in GACAN will have the same structure.}. We modify the MA implementation from \eqref{equ:ma} so that it focuses on extracting temporal features. As opposed to  \eqref{equ:ma} that focuses on the correlation between two different nodes at the same time slice, our MA captures the correlation of the same node with itself in previous time slices. 
After the attention mechanism, a fully-connected layer to learn the importance of different time intervals for the next time prediction results and a normalization layer is employed to prevent overfitting. The $K$-headed graph attention for extracting  temporal features of different time granularity, i.e., 5-minute interval features $T_{\m}$, hourly interval features $T_{\h}$, daily interval features $T_{\dy}$, and weekly interval features $T_{\w}$, can be defined as follows: For granularity $\sharp\in\{\m,\h,\dy,\w\}$, 
\begin{equation}
    \begin{aligned}
    &T_{\sharp}\coloneqq (h_{t_{\sharp}-1}^\sharp, h_{t_{\sharp}-2}^\sharp, \ldots, h_{0}^\sharp)  \\
    &h_{t}^\sharp\coloneqq \|_{k=1}^{K}\sigma\left(\sum_{i=0}^{t_\sharp-1}W_{k}^\sharp\alpha^\sharp_{t,t-is_\sharp}\mathcal{X}^{t-is_\sharp}\right)\\
    &\alpha^\sharp_{t,i}\coloneqq \frac{{\exp}(\mathrm{FC}([\mathcal{X}^{t}\|\mathcal{X}^{t-is_\sharp}]))} {\sum_{j=0}^{t_\sharp-1}{\exp}(\mathrm{FC}([\mathcal{X}^{t}\|\mathcal{X}^{t-js_\sharp}]))}
    \end{aligned}
    \label{equ:attention}
\end{equation}
where $\sigma$ is the sigmoid function that applies non-linearity, ${\|}$ denotes matrix concatenate, $\mathcal{X}^{t}\in \mathbb{R}^N$ is the input at time slice $t$,  $h_{t}^{\sharp}\in \mathbb{R}^{K\times t_\sharp \times N}$ is the feature (with time granularity indicated by $\sharp$) at time slice $t$, $\alpha_{t,i}$ is the correlation of $\mathcal{X}^t$ and $\mathcal{X}^i$, $\mathrm{FC}$ is the fully connected layer with leakyReLU activation, $W_{k}^\sharp\in \mathbb{R}^{N\times N}$ is a learnable matrix of head $k$ for time granularity $\sharp$. 
The output of the multi-attention mechanism is 
\begin{equation}\label{equ:mul_attn}
{\att}(\mathcal{X})\coloneqq {\mathrm{FC}}(T_{\m}\|T_{\h}\|T_{\dy}\|T_{\w})  
\end{equation}    

\subsection{Graph Convolution for Extracting Spatial Features}\label{sec:convolution}
Graph convolution is employed as the second layer in each ACA block. The goal  is to process network-based data and extract features in the spatial dimension. Let $\mathbf{W}$ denote the adjacency matrix of the input graph $G$ and $L\in \mathbb{R}^{N\times N}$ denote the normalized graph Laplacian as defined in \eqref{equ:laplacian}.

To avoid the expensive computation cost limitation caused by the computation of kernel $\Theta$ in graph convolution by \eqref{equ:graph_conv_1}, we apply Chebyshev polynomials approximation \cite{defferrard2016convolutional}. The method approximates the kernel $\Theta$ by Chebyshev polynomials of the diagonal matrix of eigenvalues $\Lambda$. The kernel $\Theta$ is then restricted as $\Theta(\Lambda)=\sum_{k=0}^{r-1} \theta_{k}\Theta^k$, where the kernel size $r$ denotes the maximum radius of the convolution from a central node and $\theta_k\in\mathbb{R}^{r} $ is a vector of polynomial coefficients. For input signal $X\in \mathbb{R}^{N\times Q}$, let $T_k(\cdot)$ be the Chebyshev polynomials which are recursively defined as $T_k(X) = 2X T_{k - 1} (X) – T_{k - 2} (X)$ where $T_0(X)=0$ and $T_1(X)=X$. As a result, the restricted graph convolution can then be rewritten as
\begin{equation}
\begin{aligned}
\Theta *_G X &\coloneqq \sum_{k=0}^{r-1} \theta_{k}L^{k} X \approx\sum_{k=0}^{r-1} \theta_{k} T_{k}(\tilde{L}) X
\end{aligned}
\label{equ:graph_conv_2}
\end{equation}
where the scaled Laplacian $\tilde{L}\coloneqq 2 L / \lambda_{\max}-I$ and $\lambda_{\max}$ denotes the largest eigenvalue of $L$ \cite{hammond2011wavelets}. 
With Chebyshev polynomials approximation, the cost of \eqref{equ:graph_conv_1} can be reduced to $O(K|E|)$ with $|E|$ being the number of edges as \eqref{equ:graph_conv_2} shows.

\section{Experiments}\label{sec:experiment}
\subsection{Datasets}
We evaluate the performance of our proposed model GACAN on three real-world traffic datasets \textbf{PeMSD4} where the traffic data are aggregated every 5 minutes and 29 roads are selected in our experiment, collected by the Caltrans Performance Measurement System (PeMS) \cite{chen2001freeway}, \textbf{PeMSD7} which is aggregated into minute interval from 30-second data samples, also collected by PeMS, \textbf{METR-LA} which contains traffic information collected from Los Angeles County. The time spans of the datasets are: PeMSD4 from January to March in 2018, PeMSD7 from May to June of 2018, METR-LA from March to June of 2019. Table \ref{dataset} shows the statistics of these three datasets.

\begin{table}[htbp]
\caption{Parameters of PeMSD4, PeMSD7 and METR-LA}
\begin{center}
\begin{tabular}{|c|c|c|c|}
\hline
\textbf{Dataset} & \textbf{PeMSD4} & \textbf{PeMSD7} & \textbf{METR-LA} \\
\hline
    \textbf{Location} & San Francisco & California & Los Angeles  \\ 
    \textbf{Nodes (Sensors)} & 307 & 228 & 207 \\ 
    \textbf{Start Time} & 1/1/2018 & 5/1/2018 & 3/1/2012 \\ 
    \textbf{End Time} & 3/31/2018 & 6/30/2018 & 6/30/2012 \\ 
   \textbf{Time Steps} &  25,920 & 17,568 & 25,097\\
   \textbf{Total Values} & 7,559,568 & 3,604,953 & 4,571,666\\
\hline
\end{tabular}
\label{dataset}
\end{center}
\end{table}

\subsection{Data processing}
The traffic data of three datasets are aggregated every 5 minutes, thus, every node of the road graph contains 288 data points per day. The linear interpolation method is used to solve the missing values after data cleaning problem. In addition, the input data are normalized by zero-mean method to let the average of input data be 0.
The adjacency matrix $\mathbf{W}$ of the network is computed based on the distances among stations in the traffic network, which can be formed as,
\begin{equation}
\mathbf{W}_{i j}=\begin{cases}
\exp \left(-\frac{d_{i,j}^{2}}{\tilde{\sigma}^{2}}\right), & \text{ if $i \neq j$ and  $\exp \left(-\frac{d_{i, j}^{2}}{\tilde{\sigma}^{2}}\right) \geq \epsilon$} \\
0 \quad, & \text { otherwise}
\end{cases}
\end{equation}
where $d_{i,j}$ is the distance between node $i$ and $j$. The parameters $\tilde{\sigma}^{2}=10$ and $\epsilon=0.5$ control the distribution and sparsity of matrix $\mathbb{W}$, respectively.

\subsection{Experiment Settings}
We implemented the GACAN model using the Tensorflow 2.0 framework. To eliminate atypical traffic situation. We set the number $K$ in the multi-attention to 4 as in \cite{velivckovic2017graph} and the number of terms $r$ of Chebyshev polynomial to $3$ as in \cite{kipf2016semi}. The parameter used in the leakyReLU of FC is 0.2. In our experiment the loss function is the root mean square error (RMSE) between the estimator and the ground truth and is minimized by back-propagation.

\subsection{Baselines}
We compare  GACAN with the following baselines:
\begin{enumerate}
    \item Historical Average (HA). We use the average value of the last 9 time slices to predict the next value
    \item Auto-Regressive Integrated Moving Average (ARIMA) \cite{williams2003modeling}. One of the most common statistical models for time series prediction.
    \item Long Short-Term Memory networks (LSTM) \cite{Hochreiter1997Long}. A typical RNN-based series prediction model.
    \item Graph Convolutional GRU (GCGRU) \cite{li2018adaptive}. A representative recurrent graph neural networks (RecGNN) model for traffic forecasting.
    \item Spatio-Temporal Graph Convolutional Networks (STGCN) \cite{yu2018spatio}. First adapted spatial-temporal graph convolutional Networks for traffic forecasting.
    \item Attention based Spatial-Temporal Graph Convolutional Networks (ASTGCN)\cite{guo2019attention}. A typical model adapted GAT to capture the temporal features. 
\end{enumerate}
Mean Absolute Error (MAE) and Root Mean Square Error (RMSE) are used as the evaluation metrics.
\subsection{Experimental Results}
We compare our models with the six baseline methods on PeMSD4, PeMSD7 and METR-LA. Table \ref{rescmp} demonstrate the average results of traffic speed prediction performance on the datasets PeMSD4, PeMSD7 and METR-LA over the next 15/30/60 minutes. It can be seen from table \ref{rescmp} that our GAGCN achieves the best performance in three datasets in terms of most evaluation metrics. Evidently, deep learning models generally perform better than traditional statistical and machine learning methods, due to the absence of spatial-temporal information, stationary assumption of time sequences and error accumulation. The methods based on deep learning generally obtain better prediction results than the traditional time series analysis methods due to their abilities to model nonlinear and complex traffic data and consideration among spatial-temporal data.

\begin{table*}[htbp]
\caption{Average performance comparison of different approaches on the dataset PeMSD4, PeMSD7 and METR-LA}
\begin{center}
\begin{tabular}{|c|c|c|c|c|c|c|}
\hline
\textbf{Model}&\multicolumn{2}{|c|}{\textbf{PeMSD4}}&\multicolumn{2}{|c|}{\textbf{PeMSD7}}&\multicolumn{2}{|c|}{\textbf{METR-LA}} \\
\cline{2-7}
\textbf{} & \textbf{MAE}& \textbf{RMSE}& \textbf{MAE}& \textbf{RMSE} & \textbf{MAE}& \textbf{RMSE} \\
\hline
    HA & 4.69 & 8.15 & 4.44 & 7.87 & 4.12 & 7.28\\
    ARIMA & 5.96 / 6.60 / 8.14 & 13.42 / 14.78 / 18.33 & 5.69 / 6.45 / 7.76 & 13.22 / 14.39 / 17.41 & 4.95 / 5.60 / 7.06 & 9.85 / 11.99 / 13.44 \\
    LSTM & 3.98 / 4.36 / 5.41 & 7.00 / 8.03 / 11.00 & 3.79 / 4.15 / 5.22 & 6.52 / 7.38 / 10.40 & 3.56 / 3.78 / 5.15 & 5.06 / 6.59 / 9.27  \\
    GCGRU & 2.80 / 3.59 / 5.21 & 4.46 / 6.39 / 9.48 & 2.62 / 3.52 / 4.87 & 4.61 / 6.22 / 9.24 & 2.28 / 2.97 / 4.66 & 3.54 / 5.22 / 7.41 \\
    STGCN & 2.46 / 3.17 / 4.29 & 4.22 / 6.05 / 7.96 & 2.34 / 2.99 / 4.17 & 3.99 / 5.84 / 7.87 & 2.14 / 2.75 / 3.82 & 3.39 / 4.77 / 6.54 \\
    ASTGCN & 2.24 / 2.88 / 3.84 & 3.71 / 5.32 / 7.20 & 2.21 / 2.81 / 3.79 & 3.60 / 5.21 / 7.00 & 1.86 / 2.52 / 3.57 & 2.95 / 4.12 / 6.07 \\
    \textbf{GAGCN} & {\bf 2.17 / 2.82 / 3.79} & {\bf 3.52 / 5.29 / 7.11} & {\bf 2.14 / 2.76 / 3.72} & {\bf 3.52 / 5.17 / 6.94} & {\bf 1.76 / 2.28 / 3.19} & {\bf 2.63 / 3.98 / 5.40} \\ \hline
\multicolumn{7}{l}{}\\
\multicolumn{7}{l}{15 / 30 / 60 minute forecast performance, 15 minute forecast performance of HA.}
\end{tabular}
\label{rescmp}
\end{center}
\end{table*}

\begin{figure}[!htbp]
    \centering
    \includegraphics[width=0.5\textwidth]{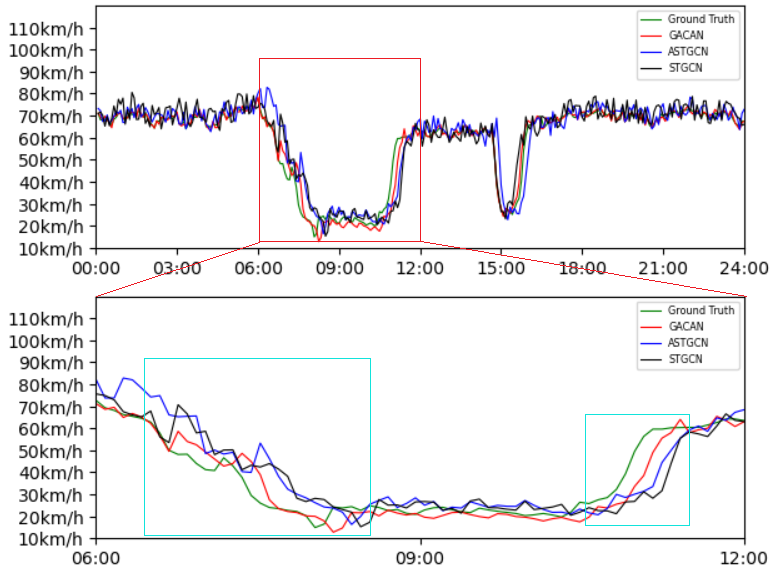}
    \caption{The prediction results of different methods on PeMSD4 show that GACAN (red lines) can produce results that are closer to the ground truth (green line).}
    \label{fig:fig4}
\end{figure}

Figure~\ref{fig:fig4} shows the one-hour-prediction performance of various methods. As shown, our model performs more accurately than the other two spatial-temporal-based modules, and captures the trend of rush hours more accurately due to the GAT units which distinguish the influence of different dates on the current date when capturing the temporal features. 

\begin{figure}[!htbp]
    \centering
    \includegraphics[width=0.49\textwidth]{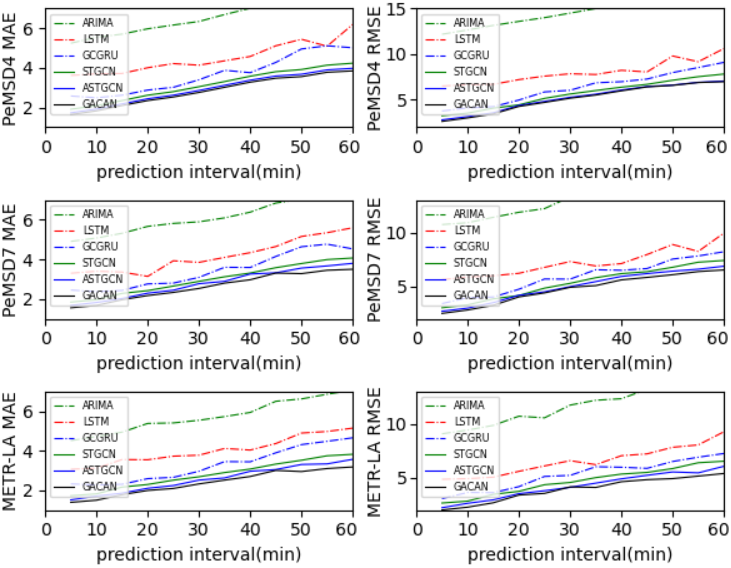}
    \caption{Performance changes of different methods as the forecasting interval increases on PeMSD4}
    \label{fig:fig5}
\end{figure}

Figure~\ref{fig:fig5} shows the prediction accuracy of various models with gradually rising prediction horizons (0-60 minutes). Overall, the  accuracy of these models decreases as the prediction horizon grows. It is clear that the performances of those models that only take temporal correlations into account, such as HA, ARIMA and LSTM, are poorer compared to models such as GCGRU, STGCN, ASTGCN and GACAN. GACAN achieves the best prediction performance in all situations. Although our model GACAN has a similar performance as ASTGCN for PeMSD4, our model needs less training time due to its simplicity.

\begin{figure}[!htbp]
    \centering
    \includegraphics[width=0.5\textwidth]{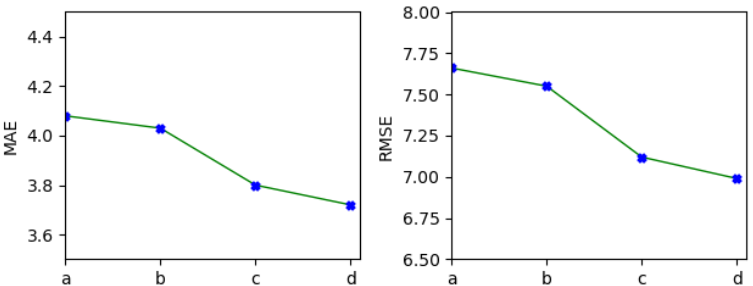}
    \caption{The one-hour-prediction results of GACAN on PeMSD4 considering time series of different granularities. 'a' stands for 5-minute; 'b' stands for 5-minute and hourly; 'c' stands for 5-minute, hourly and daily; 'd' stands for 5-minute, hourly, daily and weekly.}
    \label{fig:fig6}
\end{figure}

To validate the advantage of integrating time series of multiple time granularities, we train GACAN with different time series inputs; See figure~\ref{fig:fig6}. Clearly the model that is trained using only the original time series performs the worst. The performance improves consistently as more time granularities are introduced. The improvement is most significant when the daily time series is added, suggesting that the model aiming to capture daily traffic patterns.

\section{Conclusion and Future Work}
We propose GAGCN that combines spectral-based graph convolution with multi-attention to capture spatial-temporal features from traffic data for traffic prediction. The graph attention uses four time series as input with different time granularity: original, hourly, daily, and weekly data. Unlike ASTGCN where the four time series are processed separately, GACAN integrates the effects of all time series after each attention layer. 
Experiments on three real-world datasets show that integrating the different time series lead to an improved accuracy. The traffic speed accuracy of the proposed model is superior to other state-of-the-art models indicating that it has great potentials on exploring spatial-temporal features. 

The integration of multiple time series is an interesting technique that has the potential to produce highly accurate predictive models. 
There are still many information that could be  taken into account to improve the prediction accuracy, such as weather conditions, accident, emergencies and so on. A natural future direction is to apply data on these other influencing factors in order to improve the forecasting accuracy and adaptability. GACAN provides a general framework where these data can be integrated in the same way as the multi-time granularity data. Also, the model is a general spatial-temporal forecasting framework over graph structures, and can thus be applied to other pragmatic applications, such as knowledge graph, and recommender systems.

\bibliographystyle{plain}
\bibliography{GACAN}

\end{document}